\documentclass[12pt]{l4dc2021} 

\usepackage[utf8]{inputenc} 
\usepackage[T1]{fontenc}    
\usepackage{hyperref}       

\usepackage{booktabs}       
\usepackage{nicefrac}       
\usepackage{microtype}      
\usepackage{amsfonts}
\usepackage{bm,nicefrac}
\usepackage{gensymb}
\usepackage{mathtools}
\usepackage{siunitx}
\usepackage{multirow}
\usepackage{multicol}
\usepackage{amsmath}
\usepackage{enumitem}
\makeatletter
\def\set@curr@file#1{\def\@curr@file{#1}}
\makeatother

\title[Discovering Neural SSMs via Learning and Evolution]{Automating Discovery of Physics-Informed Neural State Space Models via Learning and Evolution}
\usepackage{times}

\author{%
 \Name{Elliott Skomski} \Email{elliott.skomski@pnnl.gov}\\
 \Name{J{\'a}n Drgo\v na} \Email{jan.drgona@pnnl.gov}\\
  \Name{Aaron Tuor} \Email{aaron.tuor@pnnl.gov}\\
 \addr Pacific Northwest National Laboratory \\
 Richland, WA, USA \\
}

\begin{document}

\maketitle

\begin{abstract}

Recent works exploring deep learning application to dynamical systems modeling have demonstrated that embedding physical priors into neural networks can yield more effective, physically-realistic, and data-efficient models. However, in the absence of complete prior knowledge of a dynamical system's physical characteristics, determining the optimal structure and optimization strategy for these models can be difficult. In this work, we explore methods for discovering neural state space dynamics models for system identification. Starting with a design space of block-oriented state space models and structured linear maps with strong physical priors, we encode these components into a model genome alongside network structure, penalty constraints, and optimization hyperparameters. Demonstrating the overall utility of the design space, we employ an asynchronous genetic search algorithm that alternates between model selection and optimization and obtains accurate physically consistent models of three physical systems: an aerodynamics body, a continuous stirred tank reactor, and a two tank interacting system.
\end{abstract}


\section{Introduction}
 Recent work has shown that given an appropriate design space, Neural Architecture Search (NAS) \citep{elsken2018neural} using evolutionary algorithms---so-called Neuroevolution~\citep{floreano2008neuroevolution}---can discover models that meet or exceed the performance of expert-designed networks for complex computer vision~\citep{liang2018evolutionary, real2019regularized}, natural language processing~\citep{so2019evolved}, and continuous control tasks~\citep{gaier2019weight}. In this work we bring together neural architecture search and works viewing neural networks from a dynamical systems perspective. We present a new NAS design space for the discovery of  accurate control-oriented systems models from a host of physics-driven neural network components. 

Several recent design spaces for NAS have composed generic neural network functional components while preserving inductive priors from successful state-of-the-art computer vision or natural language processing neural network architectures~\citep{liu2020survey}.
    On the other hand, as it applies to dynamical system identification, NAS with neuroevolutionary methods have typically used lower level components such as neurons and basic function compositions applied in black-box systems modeling without the benefit of physics-modeling based priors embedded in the design space~\citep{7974512, subudhi2011differential, subudhi2011nonlinear, 8027980, 6085706, AYALA2020105990, hatanaka2006multi, 8027980, gaier2019weight}.
Yet outside NAS research, advantageous inductive priors for dynamics modeling have recently been proposed by several key contributions which view recurrent and residual networks through the lens of traditional dynamical systems modeling. Prominent examples are guarantees on stability by constraining the deep network's linear maps using orthogonal~\citep{mhammedi2017efficient, jia2019orthogonal, wang2020orthogonal}, spectral~\citep{zhang2018stabilizing}, symplectic~\citep{haber2017stable}, anti-symmetric~\citep{chang2019antisymmetricrnn}, stochastic~\citep{tuor2020constrained}, or Schur Decomposition~\citep{kerg2019non} parametrizations. 

Our current work integrates these two orthogonal yet complementary lines of research, neural architecture search, and dynamics-inspired neural network components to expedite the discovery of effective dynamical systems models.  We present a neural block dynamics design space that encompasses an extensive range of time-invariant, block-oriented state space models. These models are built from a library of neural components using structured linear maps to imbue models with strong priors for physical modeling, encouraging stability and data efficiency. In addition to structural hyperparameters, coefficients for multi-objective loss terms penalizing constraints violations, block interactions, trajectory smoothing, and prediction error are also included in the search space. 
We evaluate our design space with two search algorithms: Random Search (RS), and an Asynchronous Genetic Algorithm (AGA). RS has no evolutionary effects, whereas AGA is designed to maximize interaction between learning and evolution, discovering performant models in a directed and expeditious manner. 
We conduct system identification experiments for both search methods on three non-autonomous systems representing a range of dynamic behavior and achieve highly accurate models with physically consistent open-loop response for each.

\section{Methods}
Our objective in this work is to develop a neural network design space for systems identification and dynamics modeling that is general enough to model a large extent of known systems and leverages inductive priors specific to dynamical systems modeling. This design space is intended to serve as both a substrate for expert built neural dynamics models and as a basis for neural architecture search to discover architectures best suited for specific systems. 
 To this end, we introduce a family of neural state space models built from components consisting of linear map parametrizations, activation functions, and neural network block components.

\subsection{Structured Linear Maps}
  Our design space includes five structured linear map parametrizations which can introduce strong inductive biases and provide guarantees suitable for a large extent of known systems models.
  For sparsity-inducing priors, we employ a Lasso variant implemented with gradient descent as described in \citet{bottou2010large}. Aditionally, we use Butterfly maps proposed by \citet{dao2019kaleidoscope}, which represent an extensive family of learnable sparse linear maps.
  We include two matrix parametrizations which enforce singular value constraints on linear maps; both methods parametrize the matrix as a product ${\bf M = U\Sigma V}$. The first method, proposed by \citet{zhang2018stabilizing}, provides orthogonal parametrizations of ${\bf U}$ and ${\bf V}$ via Householder reflections. As computationally efficient alternative, our own method initializes ${\bf U}$ and ${\bf V}$ as random orthogonal matrices and introduces a regularization term to enforce orthogonality as parameters are updated. For both factorizations, bounds $\lambda_\text{min}$ and $\lambda_\text{max}$ are placed on the nonzero elements of the diagonal matrix ${\bf \Sigma}$ where:
  \begin{equation}
    {\bf \Sigma} = \text{diag}(\lambda_\text{max} - (\lambda_\text{max} - \lambda_\text{min}) \cdot \sigma(\mathbf{p}))
  \end{equation}
  with $\mathbf{p}$ a randomly initialized vector and $\sigma$ the elementwise logistic sigmoid. 
  The final structured parametrization in the design space is the Perron-Frobenius map proposed by \citet{tuor2020constrained}, which bounds the dominant eigenvalue of the matrix to guarantee stability of the learned system and global dynamic properties such as dissipation. 

\subsection{Activation Functions}
The design space contains a library of two non-parametric and two parametric activation functions: Rectified Linear Units (ReLU), a common activation which clamps negative values to zero \citep{nair2010rectified}; Gaussian Error Linear Units (GELU), which approximates the expected value of stochastic regularization \citep{hendrycks2016gaussian}; Bendable Linear Units (BLU), which learns a continuous approximation of two piecewise linear functions \citep{godfrey2019evaluation}; and Soft Exponential (SE), which learns a function which interpolates between exponential and logarithmic functions \citep{godfrey2015continuum}. For parametric activation functions, we use independent activations at each layer to allow models to learn activations for each phase of computation.

\subsection{Neural Time-Invariant Block Dynamics Models}
The general form of state space model we consider 
is composed of a state estimator $f_o$, state transition dynamics $f$, and observation dynamics $f_y$
with learnable parameters $\theta$.
With ${\bf u}_k$ as control input at time $k$, and ${\bf y}_{k-N_p}, ..., {\bf y}_{k-1}, {\bf y}_k$ a sequence of initial observed
variables of the system the model is of the form: 

\begin{subequations}
\begin{align}
{\bf \hat{x}}_k^{\text{est}} &= f_o({\bf \hat{y}}_{k-N_p}, ..., {\bf \hat{y}}_{k} ; \theta_g) \\
{\bf \hat{x}}_{k+1}          &= f({\bf \hat{x}}_k^{\text{est}}, {\bf u}_k; \theta_x) \label{eqn:combined_input_f} \\
{\bf \hat{y}}_{k + 1}        &= f_y({\bf \hat{x}}_{k+1}; \theta_h) 
\end{align}
\end{subequations}
It is common for the state dimension of the main transition dynamics to be unknown.  From the deep learning perspective we can view ${\bf \hat{x}}$ as the hidden state of an RNN with cell function $f$, input ${\bf u}$, and output ${\bf \hat{y}}$. From a dynamics modeling perspective, when $f$ is linear, $f_o$ plays the role of a finite approximate lifting function from Koopman operator theory as suggested by \citet{yeung2019learning}. The library of block components for $f$, $f_o$, and $f_y$ consists of linear maps (LMs), multi-layer perceptrons (MLPs), residual networks (rMLPs), and recurrent neural networks (RNNs) built from the parametrized linear maps and activation functions discussed in the previous sections. An additional factor of variation is introduced in $f_o$ to account for time lag using a window of $N_p$ past observations.

Within the general class of models, $f$ may be a single neural network which takes as input the concatenated ${\bf u}$ and ${\bf \hat{x}}$ vectors. Or alternatively, lending more structure to the problem,
interactions between the the control inputs, and principle state transition dynamics are modeled as a composition component blocks so that Equation \ref{eqn:combined_input_f} becomes: 
\begin{equation}
\begin{aligned}
{\bf \hat{x}}_{k+1}  = f_x({\bf \hat{x}}_k^{\text{est}}; \theta_x)  \circ f_u({\bf u}_k; \theta_u)\\
\end{aligned}
\end{equation}
where $f_x$ and $f_u$ are drawn from the library of block components and $\circ$, is an elementwise operator modeling the influence of exogenous inputs upon the estimated system state. Three operators are considered: addition ($+$), multiplication ($\times$), and a learnable interpolation of addition and multiplication ($+/\times$) as given in Equation 5 of \citet{godfrey2015continuum}. 

The unstructured and structured model classes, which are high level options in the design space are denoted \textbf{\em black-box} and \textbf{\em block} respectively.
The block-oriented formulation allows the representation of several classes of models commonly used in systems identification. Choosing an addition operator, when both are $f_x$ and $f_u$ are linear, we have a basic linear time-invariant model. When only $f_u$ is nonlinear, we have a Hammerstein model with neural network nonlinearity. When both $f_u$ and $f_y$ are nonlinear but not $f_x$, we have a Hammerstein-Weiner model. Making all blocks nonlinear provides a general neural block nonlinear model with nonlinear state transition dynamics.

\subsection{Multi-Objective Loss}
    The state space models are trained with multi-objective loss functions informed by best practices in control-oriented data-driven system identification. Expressions for constituent objective terms are found in Equation \ref{eq:loss}.
        The principal loss function is mean squared error between predicted trajectory and ground-truth measurements over an $N$-step time horizon, $\mathcal{L}_y$ (Equation \ref{eq:pred}). The model is given a sequence of $N_p$ initial previous ground-truth measurements and future sequence of control inputs $\mathbf{U}$, then generates a series of predictions $\mathbf{\hat{y}}_1, ..., \mathbf{\hat{y}}_N$. Although not necessary, for simplicity we align the $N_p$ past measurements given to the state estimator, and the $N$-step prediction horizon so that $N_p = N$.  
        To promote alignment between the state estimator, $f_o$, and dynamics, $f$, we incorporate an additional arrival cost penalty, $\mathcal{L}_{\text{est}}$ (Equation \ref{eq:arrive}).
         We include an additional term, $\mathcal{L}_{dx}$, to ensure smooth state transitions regularizing the distance between successive states (Equation \ref{eq:smooth}). 
        We employ the penalty method such that predicted observables remain within realistic bounds, enforcing this property by defining lower and upper bounds $\underline{y}$ and $\overline{y}$, then apply inequality constraints via a loss term $\mathcal{L}^{\text{con}}_y$ (Equation \ref{eq:penalty}). 
        We can further constrain the influence of input map $f_u$ on predicted states for block-structured models, defining lower and upper bounds $\underline{f_u}$ and $\overline{f_u}$ and using the above inequality constraint formulation to create another loss term, $\mathcal{L}^\text{con}_{f_u}$. In this work, we set $\underline{y} = -0.2$, $\overline{y} = 1.2$, $\underline{f_u} = -0.02$, and $\overline{f_u} = 0.02$. 
        As shown in Equation \ref{eq:total}, the combined multi-objective loss $\mathcal{L}$ contains the terms from above weighted by factors $Q$---these loss terms may be optimally weighted with varying importance for particular systems and architectures.
        
      \begin{subequations}
\setlength{\abovedisplayskip}{0pt}
\setlength{\belowdisplayskip}{0pt}
\begin{align}
     \label{eq:pred}\mathcal{L}_y &= \frac{1}{N} \sum_{k=1}^N ||{\bf \hat{y}}_{k} - {\bf y}_{k}||^2_2    \\
       \label{eq:arrive}\mathcal{L}_{\text{est}} &= ||{\bf \hat{x}}^{\text{est}} - {\bf \hat{x}}||^2_2\\
\label{eq:smooth}\mathcal{L}_{dx} &= \frac{1}{N-1} \sum_{k=1}^{N-1} ||{\bf \hat{x}}_k - {\bf \hat{x}}_{k+1}||^2_2\\
    \label{eq:penalty}\mathcal{L}^{\text{con}}_y &= \frac{1}{N} \sum_{k=1}^N
                \bigl(
                    \text{max}(0,\:-\mathbf{\hat{y}}_k + \underline{y}) +
                    \text{max}(0,\:\mathbf{\hat{y}}_k - \overline{y})
                \bigr)\\
                \label{eq:total}\mathcal{L} &= Q_y \mathcal{L}_{y}
        + Q_\text{est} \mathcal{L}_\text{est}
        + Q_{dx} \mathcal{L}_{dx} 
        + Q^\text{con}_y \mathcal{L}^\text{con}_y
        + Q^\text{con}_{f_u} \mathcal{L}^\text{con}_{f_u}
        + Q_\text{reg} \mathcal{L}_\text{reg}
\end{align}\label{eq:loss}
\end{subequations}  
    
\begin{figure}[t]
        \centering
        \includegraphics[trim=75 0 75 0,clip,width=\linewidth]{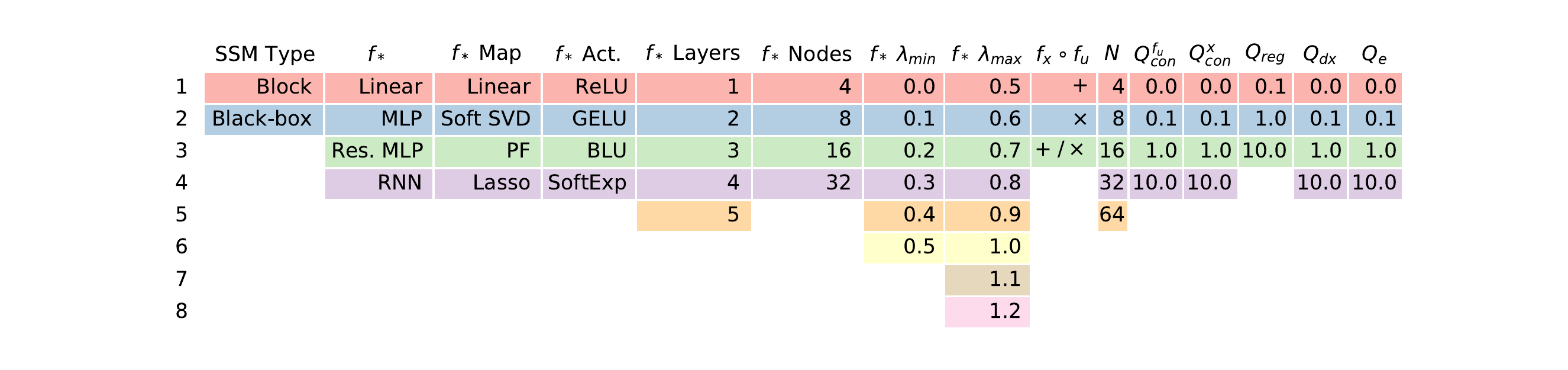}
    \caption{The SSM genome.}
    \label{fig:genome}
\end{figure}

\subsection{State Space Model Genome}
        Figure \ref{fig:genome} illustrates the complete space of model configurations possible in our proposed neural dynamics design space. Columns indicate the attributes (genes) of constituent components while rows indicate the possible values each attribute can take. Columns with $f_*$ represent attributes that model components $f_x$, $f_u$, $f_y$, and $f_o$ share but whose values may vary across components. For the purposes of our experiments, we represent models sampled from the search space as a vector of traits, each with a discrete set of possible values. These sets are represented as ordered ring buffers which allows wrapping of position-wise mutations selecting the nearest higher or lower value for attributes with a natural ordering. 
        
        In addition to the full granularity of the design space with all potential combinations of attribute values available (dubbed \textbf{\em XL}), we introduce a restricted design space (dubbed \textbf{\em standard}) that captures typical heuristics commonly found in expert designed models. The standard design space couples the linear parametrizations and activation functions of all block components $f_*$ in the SSM. It further imposes some additional hierarchy on the design of discovered architectures through the addition of SSM types Hammerstein, Hammerstein-Weiner, Linear, and Block Nonlinear. With these additional values the attribute SSM Type completely determines which block components are linear (see Table \ref{tab:block}) for the standard design space. The standard genome additionally omits $f_x \circ f_u$ from the search, using addition to model state and input component interactions by default. Following a common pattern in neural network architecture design practice, values for number of layers, number of nodes, activation function, and linear map parametrization are the same across block components in the standard design space. While still a vast search space, these restrictions reduce the total number of possible architectures by several orders of magnitude from $\sim$3 trillion for the XL design space to $\sim$3.5 billion for the standard design space. 
        
\begin{table}[t]
    \centering
    \begin{tabular}{l|cccc}
        \toprule
      {}  & \multicolumn{4}{c}{Model class}\\
       Block  & Block Nonlinear & Hammerstein-Wiener & Hammerstein & Linear\\
        \midrule
        $\mathbf{f}_x$ &N &Y &Y & Y\\
        $\mathbf{f}_u$ &N &N &N & Y\\
        $\mathbf{f}_y$ &Y &N &Y & Y\\
        \bottomrule
    \end{tabular} \\
    \caption{Linear components for standard design space block-oriented model classes.}
    \label{tab:block}
    \vspace{-10pt}
\end{table}
        
          
        \subsection{Asynchronous Genetic Search}
            We introduce an asynchronous genetic algorithm designed to maximize utilization of a fixed computation budget while searching the design space in a directed manner. This is accomplished by maintaining a fixed number of actively training individuals---initially generated at random---and at fixed duration intervals (5 minutes for our experiments) spawning new individuals as models finish training. 
            Models are ranked and selected according to best-observed $n$-step prediction mean squared error $\mathcal{L}_y$ (Equation \ref{eq:pred}) on the validation set.
        
             
        
            The algorithm is initialized with a random population of state space models from the design space. The number of new state space models dispatched during the periodic spawning phases equals the number of models which have terminated training since the last spawning, thereby effectively leveraging but not exceeding a fixed computational budget.
              
            
            New population members are generated via one of three operators; random, mutation, or crossover. The random operator selects all traits at random. The mutation operator randomly selects a trait and randomly steps to the next higher or lower value according to the trait's prescribed or natural ordering. The crossover operator implements crossover weighted by fitness, taking two individuals $A$ and $B$ and for each trait randomly selecting to use $A$'s trait with probability $\text{fitness}(A)/(\text{fitness}(A) + \text{fitness}(B))$.
            
            
            The proportion of each generation's new births via mutation versus crossover is a hyperparameter: mutation probability $p^{\text{mut}}$ is given, and crossover probability $p^{\text{cross}}$ is derived simply as $1 - p^{\text{mut}}$. Further, in order to promote more exploration at the beginning of search, each new generation may spawn a randomly-selected model with probability $p^{\text{birth}}_i$, which is annealed by a constant factor $0<k<1$ prior to each iteration, $i$, so that $p^{\text{birth}}_{i+1} = k p^{\text{birth}}_{i}$.
            
            We use random search as an alternative method for exploring the design space. Random search selects options for each position in the model vector representation at random without the fitness based guidance of genetic operators.


\section{Experiments}
To measure the efficacy of our dynamics model design space, we perform model search using the two search algorithms and three datasets. 
    \subsection{Model Training}
        Models are trained via full-batch gradient descent with the AdamW optimizer \citep{loshchilov2017decoupled} for a fixed number of epochs and a learning rate of $\num{2e-3}$. Early stopping is used to terminate training when a model has not improved on its best validation set performance for a set number of epochs. We train models for 1,000 epochs at most, and allow models 100 epochs of training without improvement before termination.
        
        We initialize each search algorithm with 50 individuals, maintain a fixed-size pool of 50 active individuals, and check the active population and spawn any new individuals every 5 minutes. For the AGA, we let initial random birth probability $p^{\text{birth}}_0 = 1$ with annealing rate $k=0.5$, and we let $p^{\text{mut}} = 0.2$ and $p^{\text{cross}} = 0.8$.
    
    \subsection{Datasets}
        We evaluate our design spaces using three non-autonomous systems with different properties:
        \begin{itemize}[noitemsep]
            \item {\bf Aerodynamics}: models $y$ and $z$ acceleration and angular velocity in all dimensions of an aerodynamic body using ten inputs \citep{aero}.
            \item {\bf CSTR}: models temperature and chemical concentration of a non-adiabatic continuous stirred tank reactor using three inputs \citep{cstr, cstr_mathworks}.
            \item {\bf Two Tank}: models water levels in two tanks governed by inputs for pump speed and valve opening \citep{tank}.
        \end{itemize}
\section{Results and Analysis}


\begin{table}[t]
    \centering
        { \scriptsize
        \begin{tabular}{llcccc}
            \toprule
            Dataset                       &      &                AGA &             Random &             AGA XL &     Random XL \\
            \midrule                                                               
            \multirow{2}{*}{Aero}         & Val. & \num{3.16e-3}      & \num{4.18e-3}      & \bf{\num{1.89e-3}} & \num{1.35e-2} \\
            {}                            & Test & \num{1.23e-2}      & \num{1.04e-2}      & \bf{\num{6.87e-4}} & \num{1.44e-2} \\ \midrule
            \multirow{2}{*}{CSTR}         & Val. & \num{7.00e-3}      & \bf{\num{6.76e-3}} & \num{8.30e-3}      & \num{7.66e-3} \\
            {}                            & Test & \num{8.32e-3}      & \bf{\num{8.00e-3}} & \num{9.14e-3}      & \num{1.12e-2} \\ \midrule
            \multirow{2}{*}{Two Tank}     & Val. & \bf{\num{3.37e-4}} & \num{5.45e-4}      & \num{4.01e-4}      & \num{3.29e-3} \\
            {}                            & Test & \bf{\num{1.00e-3}} & \num{2.21e-3}      & \num{2.56e-3}      & \num{1.11e-2} \\
            \bottomrule
        \end{tabular}
        }
    \caption{Validation and test set open-loop MSE for each dataset and algorithm's best model.}
    \label{tab:results}
\end{table}

\begin{figure}[t]
    \centering
    \subfigure[Aerodynamics]{
        \includegraphics[width=0.31\textwidth]{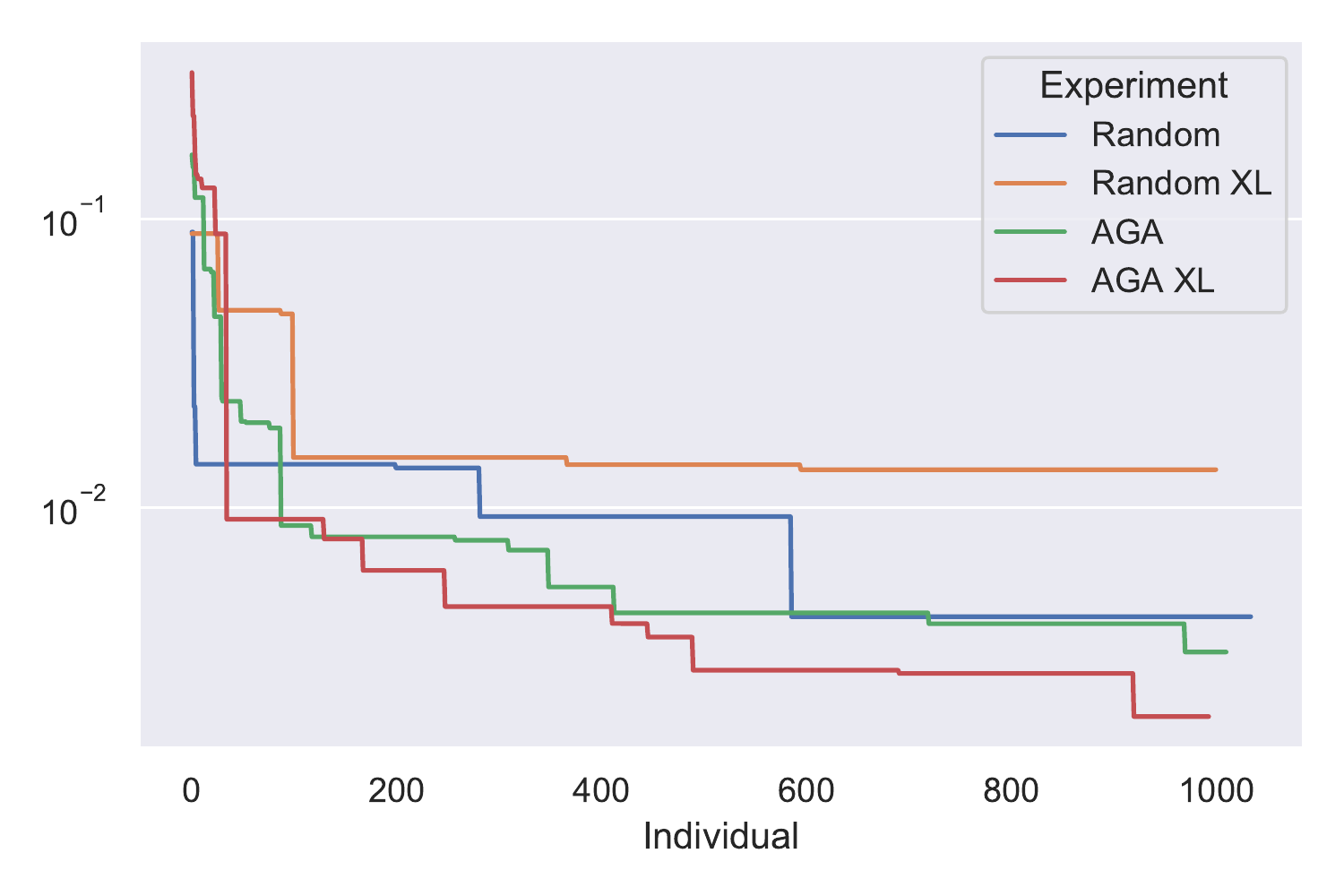}
    }
    \subfigure[CSTR]{
        \includegraphics[width=0.31\textwidth]{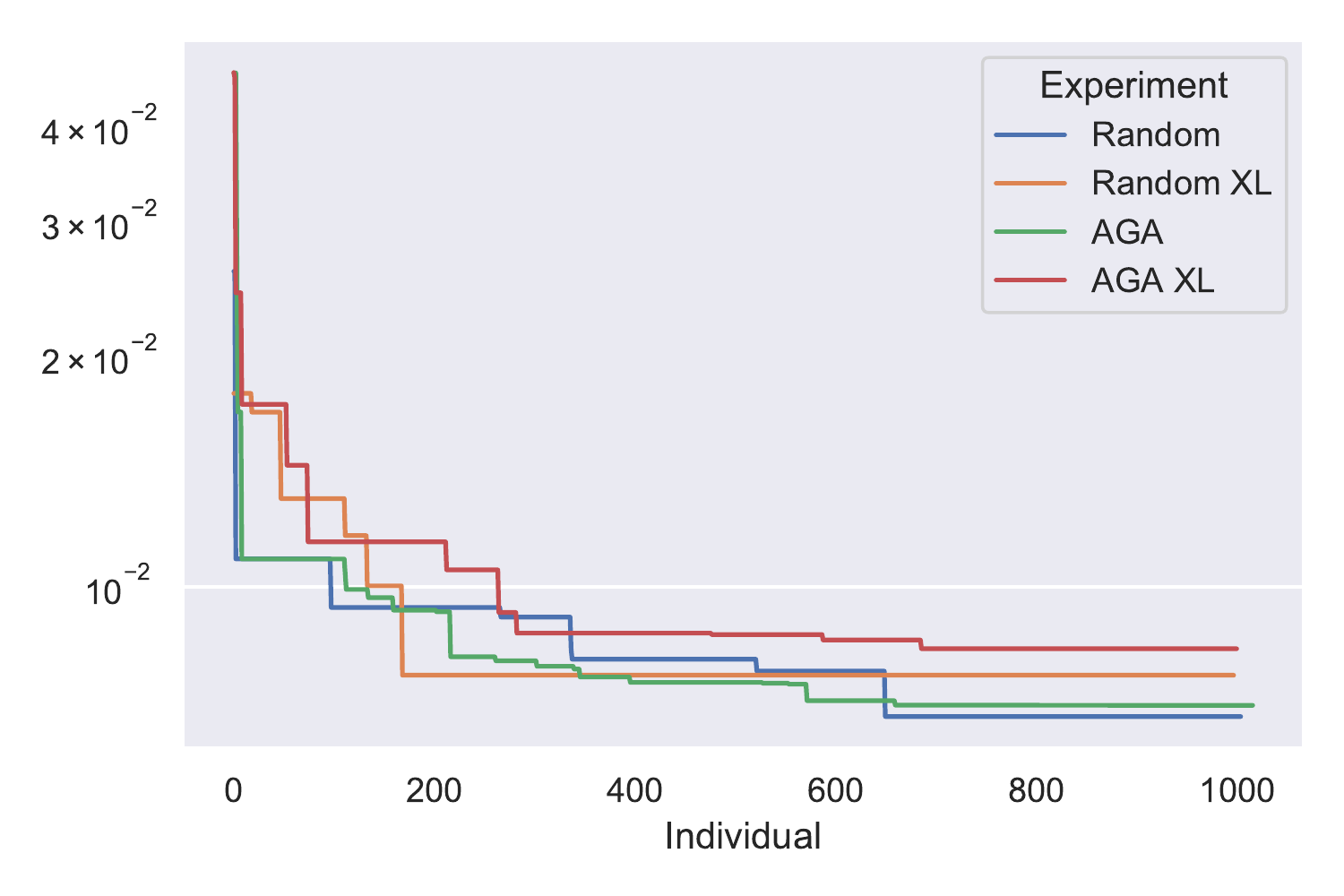}
    }
    \subfigure[Two Tank]{
        \includegraphics[width=0.31\textwidth]{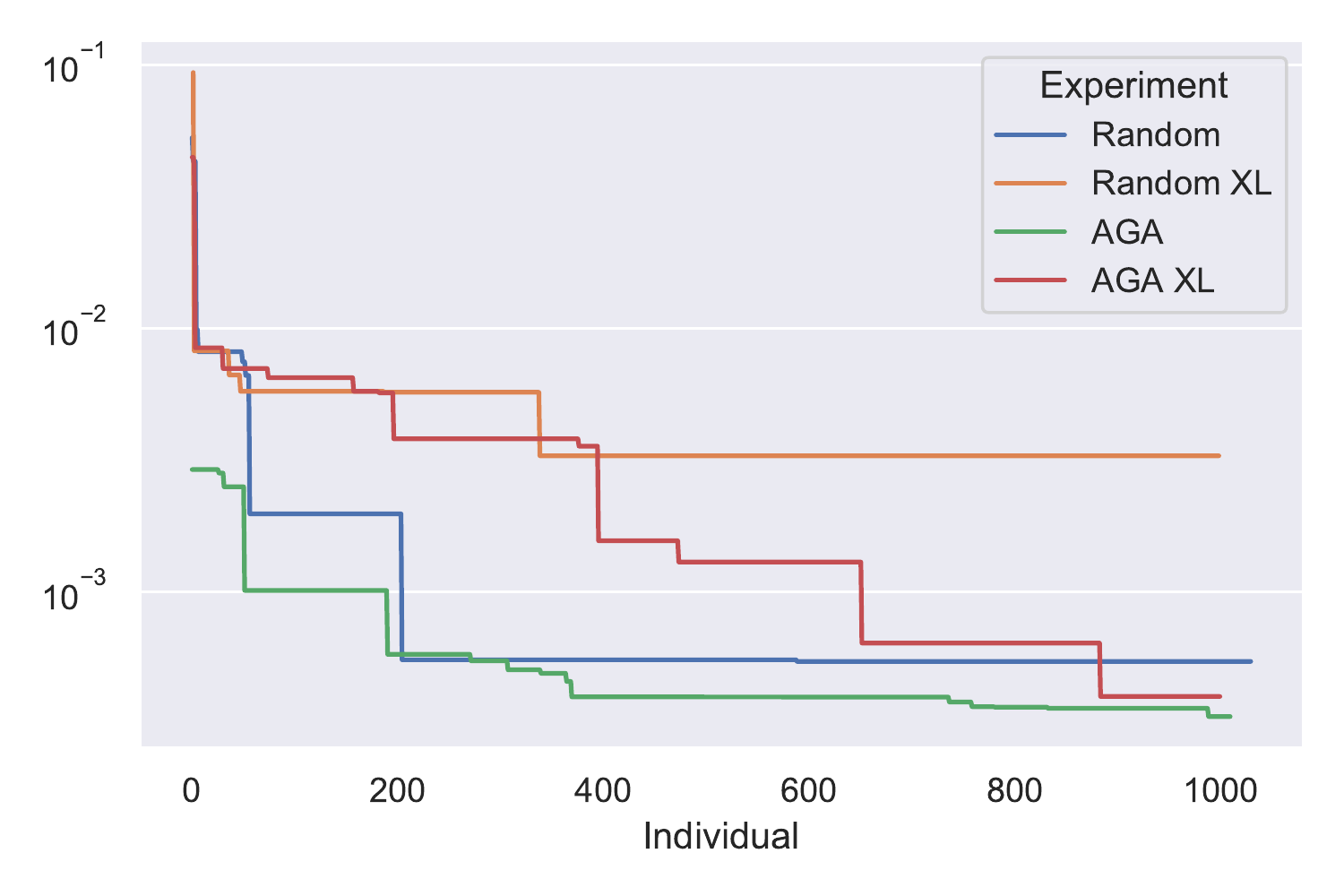}
    }
    \caption{Best open-loop validation set MSE as optimization progresses for each search algorithm, model genome, and dataset.}
    \label{fig:alg_progress}
\end{figure}

Table \ref{tab:results} lists open-loop validation and test set MSE for models obtained by each search algorithm, and Table \ref{tab:hparams} gives the attribute values for each of these models. 
For the Aerodynamics and Two Tank datasets, AGA discovered models from the XL genome that outpaced those found by RS from the same genome, with a considerable performance gap emerging as optimization progressed. These results underscore the AGA's robustness with high-dimensional search spaces. A similar result can be seen for the Two Tank datasets with the standard model genome, though the performance gap between AGA and RS isn't as pronounced. Although the Aerodynamics model obtained from the standard genome outperformed the model obtained by RS, it didn't quite generalize to the test set as effectively as the latter, a possible indication of the AGA's potential to overfit the validation set. For CSTR, RS prevailed over AGA for the standard genome in both validation and test set MSE; however, AGA search over the XL genome obtained a model with better generalization to the test set than the model found via RS despite lower validation set performance. By all accounts, AGA search was able to find models that yielded performance competitive or superior to those found by RS.

The XL genome results noticeably demonstrate the efficacy of our AGA implementation when performing directed searches in high-dimensional optimization landscapes. For the Aerodynamics and Two Tank datasets, AGA is able to pull ahead of RS and converge to more performant models. These results further suggest that decoupling the structure of model components in neural SSMs can improve performance on certain systems---the Aerodynamics model in particular benefits from this decoupling, though other systems seem to perform well without this decoupling.

\begin{table}[t]
\resizebox{\textwidth}{!}{
\begin{tabular}{lllllllrrrrrrrrrr}
\toprule
Dataset & Algorithm &     SSM Type & State Est. & Linear Map & Nonlin. Map & Activation & Layers & Nodes &  $\lambda_\text{min}$ &  $\lambda_\text{max}$ &  $N$ &  $Q^\text{con}_{fu}$ &  $Q^\text{con}_x$ &  $Q_\text{reg}$ &  $Q_{dx}$ & $Q_{f_o}$ \\
\midrule
Aero & AGA     & Blk. Nonlin. & Res. MLP & Soft SVD &           RNN &        BLU &       2 &       32 &        0.0 &        1.2 &      64 &        0.0 &      0.1 &    0.1 &   0.0 &   1.0 \\
CSTR & RS      & HW           & RNN      & Linear   &           RNN &       GELU &       3 &       32 &        --- &        --- &      64 &       10.0 &      0.0 &    1.0 &  10.0 &   0.0 \\
Two Tank & AGA &  Hammerstein & Linear   & Linear   &           RNN &       GELU &       2 &       32 &        --- &        --- &       8 &        1.0 &      0.0 &    0.1 &   1.0 &  10.0 \\
\bottomrule

\end{tabular}
}

\vspace{8pt}

\resizebox{\textwidth}{!}{
\begin{tabular}{lll|llllrrrr|crrrrrr}
\toprule
Dataset               & Algorithm &  Model & $f_*$ & Nonlin. Map & Linear Map & Activation & Layers & Nodes & $\lambda_\text{min}$ & $\lambda_\text{max}$ & $f_x \circ f_u$ &  $N$ &  $Q^\text{con}_{fu}$ &  $Q^\text{con}_x$ & $Q_\text{reg}$ & $Q_{dx}$ & $Q_{f_o}$ \\
\midrule
\multirow{4}{*}{Aero} & \multirow{4}{*}{AGA} & \multirow{4}{*}{Block} & $f_x$ &   rMLP &    Lasso &     ReLU &       1 &     32 &   --- &  --- & \multirow{4}{*}{$+/\times$} & \multirow{4}{*}{4} & \multirow{4}{*}{0.0} & \multirow{4}{*}{0.0} &   \multirow{4}{*}{10.0} &   \multirow{4}{*}{1.0} &  \multirow{4}{*}{0.1} \\
                              &                         &           & $f_u$ & Linear & Soft SVD &  SoftExp &       4 &      4 &   0.0 &  0.8 \\
                              &                         &           & $f_y$ &   rMLP &    Lasso &     ReLU &       1 &     32 &   --- &  --- \\
                              &                         &           & $f_o$ &    RNN &    Lasso &      BLU &       2 &      8 &   --- &  --- \\
\midrule
\multirow{3}{*}{CSTR} & \multirow{3}{*}{RS} & \multirow{3}{*}{Black-box} & $f_{xu}$  & rMLP &  Soft SVD &   ReLU &     5 &     32 &   0.5 &  0.9    &  \multirow{3}{*}{N/A} &     \multirow{3}{*}{16} &       \multirow{3}{*}{---} &      \multirow{3}{*}{0.1} &   \multirow{3}{*}{10.0} &  \multirow{3}{*}{10.0} &  \multirow{3}{*}{0.1} \\
                      &                            &                            & $f_y$ &    MLP   &  Soft SVD &  BLU &        2 &      4 &   0.3 & 1.2 \\ 
                      &                            &                            & $f_o$ & Linear   &        PF & GELU &        5 &      4 &   0.0 & 1.2 \\
\midrule
\multirow{3}{*}{Two Tank} & \multirow{3}{*}{AGA} & \multirow{3}{*}{Black-box} & $f_{xu}$ &   rMLP &   Linear &   GELU &      1 &     16 &   --- & --- &  \multirow{3}{*}{N/A} &   \multirow{3}{*}{8} &        \multirow{3}{*}{---} &     \multirow{3}{*}{10.0} &   \multirow{3}{*}{---} &   \multirow{3}{*}{0.1} &  \multirow{3}{*}{0.1} \\
                      &                            &                             & $f_y$    &    MLP &   Linear &    BLU &       1 &     16 & --- & --- \\
                      &                            &                             & $f_o$    & Linear &   Linear &    BLU &       5 &     32 & --- & --- \\
                          
\bottomrule
\end{tabular}
}
\caption{Attributes of best-observed models according to validation set MSE, obtained from standard (top) and XL (bottom) SSM genomes.}
\label{tab:hparams}
\end{table}

\begin{figure}[t]
    \centering
    \includegraphics[width=\textwidth]{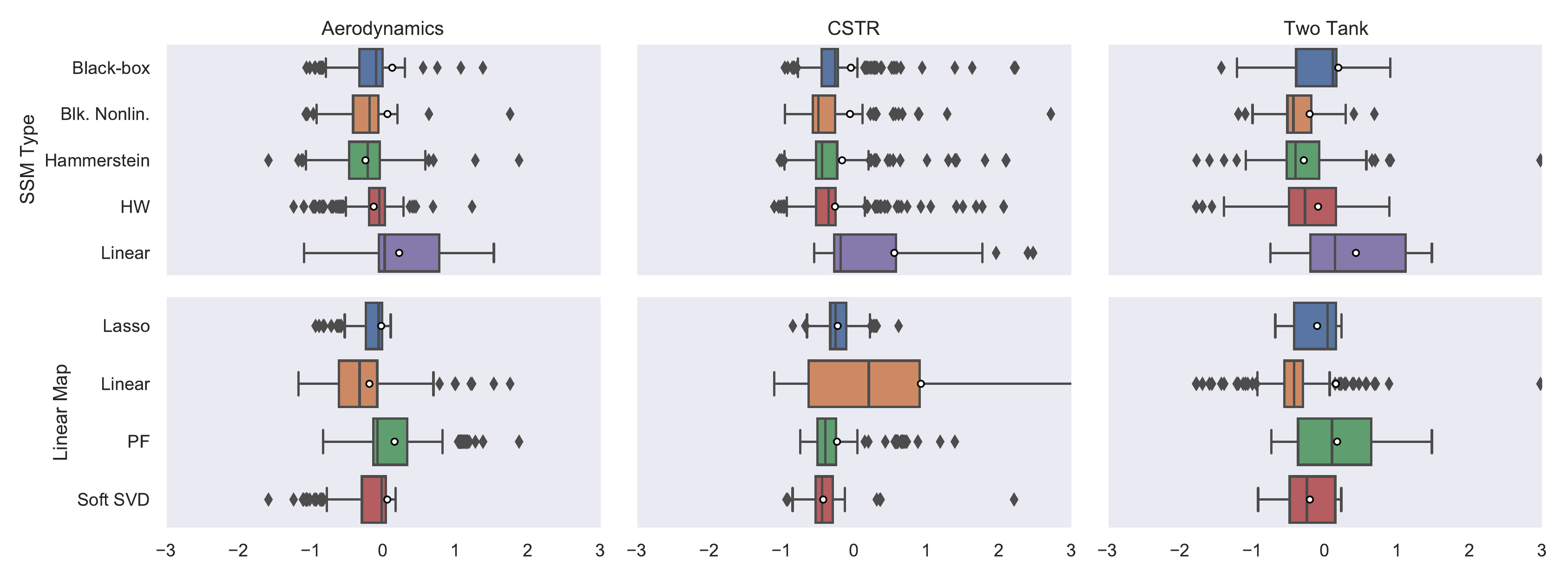}
    \caption{Boxplots indicating distributions of deviation from average log-scaled test set loss for SSM type and linear map choices used in all random search runs over the standard model genome for each dataset. White circles indicate mean of deviations.}
    \label{fig:comparisons}
\end{figure}

\begin{figure}[t]
    \centering
    \subfigure[Aerodynamics (AGA XL)]{
        \includegraphics[trim=0 0 40 40, clip, width=0.31\linewidth]{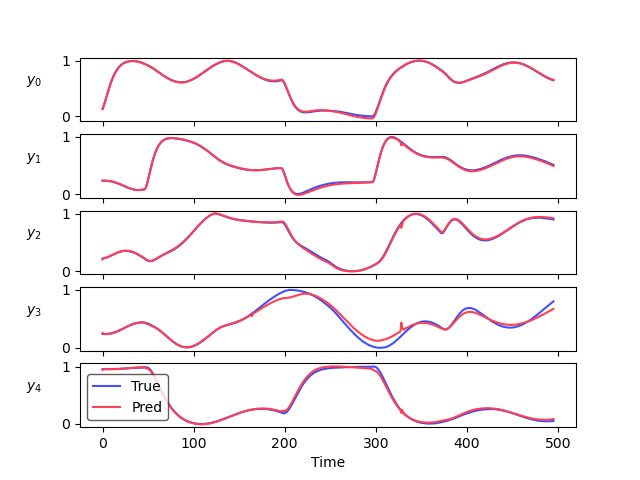}
    }
    \subfigure[CSTR (Random)]{
        \includegraphics[trim=0 0 40 40, clip, width=0.31\linewidth]{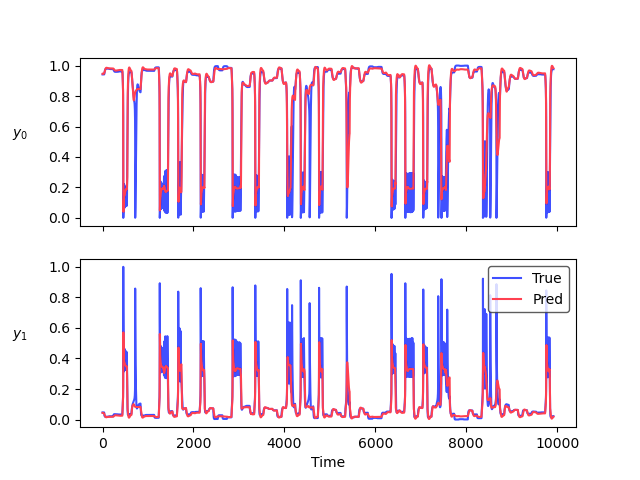}
    }
    \subfigure[Two Tank (AGA)]{
        \includegraphics[trim=0 0 40 40, clip, width=0.31\linewidth]{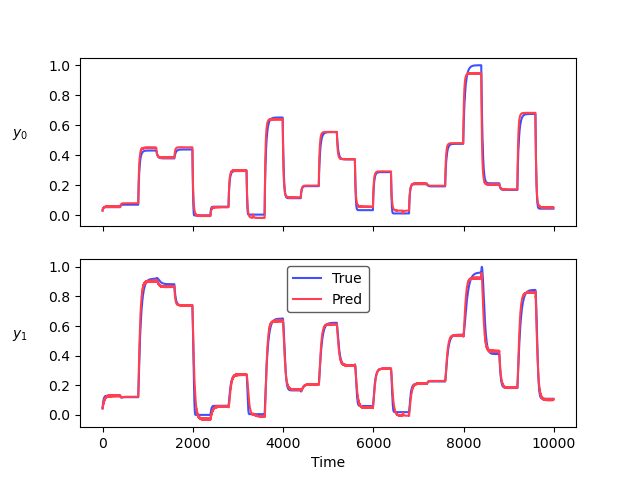}
    }
    \caption{Open-loop traces for each system's best-observed model, with search algorithm and space indicated in parentheses.}
    \label{fig:best_traces}
\end{figure}

To assess the effectiveness of structured architectures and linear maps for neural SSMs, Figure \ref{fig:comparisons} shows the distribution of deviation from the average log-scaled test set loss for models obtained by RS over the standard model genome, grouped by SSM type and linear map choices. For brevity, we focus our analysis on these attributes and their behavior in the standard genome.

One might expect that each system would converge toward a structure that lends itself to the true physical properties of the system; however, we find that a broad set of structural configurations can produce good models overall. Looking at SSM types, it appears that block-structured SSMs generally perform better than linear or black-box SSMs for all systems, though linear and black-box can be competitive at their best. Models with purely linear components can perform well, but exhibit high variance in performance---a reasonable result since the systems being modeled are nonlinear. As for linear maps, adding structure can be effective toward producing consistent models with lower variance in performance. However, this structure is not always effective for all systems---often linear maps without structural priors yield models that perform well. For certain systems, however, models with structured maps on average perform better than the average over all linear maps and may exceed the performance of unstructured maps with further tuning. It remains to be seen whether more careful tuning of constraints and loss coefficients will give better performance to the structured maps that rely on these attributes.

\section{Conclusion}
We present design space variations for neural SSMs, which include a rich selection of structured components and optimization constraints for the automated discovery of performant dynamical system identification models. We evaluate our two design spaces with two search algorithms---random and asynchronous genetic search---and provide an analysis of the architectures discovered for models trained and evaluated on three non-autonomous systems. We find that adding block structure and nonlinearity to neural SSMs can yield a greater likelihood of obtaining good models. Blackbox and/or linear variations can also perform well, though with larger variance across runs.

For future work, we would like to extend our analysis to how AGA parameters affect optimization progress and results, as well as a larger selection of structural configurations. We also plan to continue exploring the properties of our genetic algorithm. Thus far, our analysis has revealed that the asynchronicity of our genetic algorithm results in a complex interaction between learning and evolution. We hope to examine these properties in greater detail and find ways to exploit them to improve the AGA's efficacy. Additionally, the AGA currently uses a naive selection approach, which may limit the diversity of selected models. We believe this could be improved using techniques such as non-dominated sorting with multiple objectives or tournament selection. Finally, we plan to extend the AGA to evolve network topology at a lower level.

\section{Acknowledgments}
This research was supported by the Laboratory Directed Research and Development (LDRD) at Pacific Northwest National Laboratory (PNNL). 
PNNL is a multi-program national laboratory operated for the U.S. Department of Energy (DOE) by Battelle Memorial Institute under Contract No. DE-AC05-76RL0-1830.

\bibliography{references}

\end{document}